\pdfoutput=1 
\documentclass[conference]{IEEEtran}
\def\BibTeX{{\rm B\kern-.05em{\sc i\kern-.025em b}\kern-.08em
    T\kern-.1667em\lower.7ex\hbox{E}\kern-.125emX}}
\usepackage{glossaries}
\usepackage{graphicx}
\usepackage[dvipsnames]{xcolor}
\usepackage{todonotes}
\usepackage{cite}
\usepackage{amsmath, amsfonts}
\usepackage[capitalise]{cleveref}
\usepackage{siunitx}
\usepackage{makecell}
\usepackage{threeparttable}
\usepackage{lipsum}

\newcommand{\oursTable}[1]{{\color{RoyalBlue}\textit{\textbf{#1}}}}

\newacronym{pe}{PE}{Printed Electronics}
\newacronym{nre}{NRE}{Non-Recurring Engineering}
\newacronym{ml}{ML}{Machine Learning}
\newacronym{mlp}{MLP}{Multi-Layer Perceptron}
\newacronym{svm}{SVM}{Support Vector Machine}
\newacronym{mac}{MAC}{Multiply-Accumulate}
\newacronym{ovo}{OvO}{One-vs-One}
\newacronym{ovr}{OvR}{One-vs-Rest}
\newacronym{sv}{SV}{Support Vector}
\newacronym{rom}{ROM}{Read-Only Memory}
\newacronym{mux}{MUX}{Multiplexer}
\newacronym{adc}{ADC}{Analog-to-Digital Converter}

\makeatletter
\newcommand*\titleheader[1]{\gdef\@titleheader{#1}}
\AtBeginDocument{%
  \let\st@red@title\@title
  \def\@title{%
    \bgroup\normalfont\normalsize\centering\@titleheader\par\egroup
    \vskip1ex\st@red@title}
}
\makeatother

\author{\IEEEauthorblockN{
Spyridon~Besias,
Ilias~Sertaridis,
Florentia~Afentaki,
Konstantinos~Balaskas
and~Georgios~Zervakis}
\IEEEauthorblockA{Department of Computer Engineering \& Informatics, University of Patras, Patras, Greece}
\IEEEauthorblockA{\{st1072524, st1072480, afentaki, kompalas, zervakis\}@ceid.upatras.gr\vspace{-2ex}}}
\title{\vspace{-8pt}Late Breaking Results: Energy-Efficient Printed Machine Learning Classifiers with Sequential SVMs}
\titleheader{\vspace{-20pt}To appear at the  Design, Automation and Test in Europe Conference (DATE'25), March 31 - April 2, 2025.}

\begin{document}
\bstctlcite{IEEEexample:BSTcontrol} 
\maketitle

\begin{abstract}
\gls{pe} provide a mechanically flexible and cost-effective solution for machine learning (ML) circuits, compared to silicon-based technologies.
However, due to large feature sizes, printed classifiers are limited by high power, area, and energy overheads, which restricts the realization of battery-powered systems.
In this work, we design sequential printed bespoke \gls{svm} circuits that adhere to the power constraints of existing printed batteries while minimizing energy consumption, thereby boosting battery life.
Our results show 6.5x energy savings while maintaining higher accuracy compared to the state of the art.
\end{abstract}

\begin{IEEEkeywords}
Machine Learning, Support Vector Machines, Printed Electronics
\end{IEEEkeywords}

\glsresetall

\section{Introduction}
\label{sec:introduction}
\gls{pe} have emerged as a transformative technology designed to address the limitations of traditional silicon-based systems~\cite{Bleier:ISCA:2020:printedmicro}.
Printed devices offer mechanical flexibility, conformality, non-toxicity and ultra-low manufacturing and \gls{nre} costs.
However, \gls{pe} come with large feature sizes, leading to strict power and area constraints, thus making the realization of complex datapaths--such as \gls{ml} algorithms--challenging.
Leveraging the low \gls{nre} and fabrication costs of \gls{pe}, unconventional computing paradigms such as bespoke (i.e., fully customized circuits with hardwired values), and approximate computing have been exploited towards the realization of battery-powered printed \gls{ml} circuits~\cite{Mubarik:MICRO:2020:printedml,Armeniakos:TCAD2023:cross, Armeniakos:TC2023:codesign}.
However, the state of the art mainly targets reducing area overheads, neglecting energy efficiency, which is crucial for extending battery life in printed applications.

In this work, we address these limitations and propose a \gls{ml} classifier design that combines state-of-the-art accuracy with utmost energy efficiency.
We focus on \glspl{svm} due to their effectiveness in classification tasks relevant to \gls{pe} applications, and design sequential printed \glspl{svm} that compute one support vector per cycle, compressing the required compute engine and minimizing energy requirements.
Moreover, the \gls{ovr} algorithm is selected to minimize hardware requirements associated with support vectors storage.
Our SVMs achieve $6.5\times$ average energy reduction compared to state-of-the-art approaches while achieving higher accuracy.

\section{Proposed Printed Sequential SVMs}
\label{sec:svms}
\begin{figure}
    \centering
    \includegraphics[width=.95\linewidth]{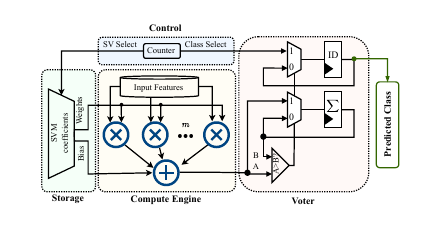}
    \vspace{-2ex}
    \caption{Abstract overview of our proposed sequential \gls{svm} circuit design.}
    \vspace{-3ex}
    \label{fig:svm_circuit}
\end{figure}
\textbf{Algorithmic Implementation:}
\glspl{svm} are supervised learning algorithms that classify data by identifying an optimal hyperplane in a high-dimensional space.
Effective for small, high-dimensional datasets, they are robust against overfitting by focusing on critical data points (i.e., support vectors) which define decision boundaries, based on which a series of classifiers can determine the predicted class.
We employ linear kernels in our \glspl{svm}, due to their simplicity and reduced hardware complexity.
Any linear classifier with $m$ weights ($w_i$) and bias $b$ computes the following weighted sum: $y = \sum_{i=1}^{m} w_i x_i + b$,
where $x_i$ are the input activations.
Targeting multi-class classification, \glspl{svm} are built upon two mainstream algorithms: \gls{ovo} and \gls{ovr}. 
\gls{ovo} creates $\frac{n(n-1)}{2}$ binary classifiers (for $n$ classes), each trained to distinguish between a specific pair of classes.
In contrast, $n$ classifiers are required in \gls{ovr} for separating each class from all others.
Since \gls{ovr} needs fewer support vectors, it is used in our implementation, targeting reduced hardware overheads.
We train our \glspl{svm} with low-precision inputs and post-training, we quantize the \gls{svm} weights and biases to the lowest precision that can retain acceptable accuracy.

\textbf{Hardware Implementation:}
We implement our trained \glspl{svm} as sequential architectures, targeting high energy efficiency alongside low area and power consumption.
An overview of our proposed sequential \gls{svm} circuits is presented in \cref{fig:svm_circuit}.
They comprise four major components: control, storage, compute engine, and voter.
Orchestrated by our control circuitry, the classification output is produced over multiple cycles, aggregating to $n$ total.
In each cycle, a selected support vector is fetched from storage, for its associated weighted sum to be calculated by our compute engine.
A log$_2(n)$-bit counter is employed for control, responsible for accessing the stored support vectors and terminating the multi-cycle process once all classifiers have been computed.
We opt for bespoke \gls{mux}-based storage units, i.e., the inputs of the \gls{mux} (excluding the control signal) are hardwired to the parameters of the support vectors.
This is made feasible by the low costs in \gls{pe}.
We also evaluated a crossbar-based \gls{rom} alternative~\cite{Bleier:ISCA:2020:printedmicro}; however for the required storage size, crossbars prove more costly, mainly due to the need for printed \glspl{adc}.
The control counter produces the select signal of the \gls{mux}.
The above showcase the two-fold advantage of using the \gls{ovr} algorithm:
compared to \gls{ovo}, which is used in the state of the art, fewer support vectors need to be stored, and less complicated control signals are needed, thus minimizing overheads at both the control and storage components. 
The entire SVM computation is folded over one compute engine,
which computes the weighted sum for each support vector fetched from the \gls{mux}.
Our engine instantiates $m$ multipliers and a multi-operand adder, thus computing one classifier per cycle and significantly reducing the hardware resources compared to fully parallel architectures~\cite{Mubarik:MICRO:2020:printedml,Armeniakos:TCAD2023:cross}, where dedicated hardware per coefficient is required.
The output of our compute engine is sent to our voter, which tracks the classifier (i.e., counter value) with the highest score (i.e., weighted sum).
Hence, our voter--essentially a sequential \texttt{argmax}--requires only two registers (for score and classifier id) and a single comparator, as finding the maximum score involves one comparison per cycle between the current and stored scores.

\section{Evaluation \& Results}
\label{sec:evaluation}
\textbf{Experimental Setup:}
We evaluate our printed sequential \glspl{svm} over five datasets from the UCI \gls{ml} repository~\cite{Dua:2019:uci}.
Synopsys Design Compiler and PrimeTime are used for hardware evaluation with the EGFET PDK~\cite{Bleier:ISCA:2020:printedmicro}, targeting frequency in the \si{\hertz} range, typical for printed applications~\cite{Bleier:ISCA:2020:printedmicro}.
Our SVMs are trained with normalized inputs to $[0,1]$ and a random $80$\%/$20$\% split for training/testing data subsets.
Accuracy is reported on the test dataset.
For comparisons, we consider the printed fully-parallel \glspl{svm}~\cite{Mubarik:MICRO:2020:printedml,Armeniakos:TCAD2023:cross} and \glspl{mlp}~\cite{Armeniakos:TC2023:codesign}, since they achieve high accuracy.

\begin{table}[t!]
\setlength\tabcolsep{3pt}
\caption{Hardware evaluation of our proposed sequential \acrshortpl{svm} and comparison with state-of-the-art techniques~\cite{Mubarik:MICRO:2020:printedml, Armeniakos:TCAD2023:cross, Armeniakos:TC2023:codesign}.}
\label{tab:results}
\centering
\renewcommand{\arraystretch}{1.1}
\begin{threeparttable}

\begin{tabular}{ll|cccccc}

\hline
    \textbf{Dataset$^\dagger$} & \textbf{Model} & 
    {\thead{\textbf{Acc.}\\ (\%)}} & 
    {\thead{\textbf{Area} \\ (\si{\square\centi\meter})}} & 
    {\thead{\textbf{Power} \\ (\si{\milli\watt})}} & 
    {\thead{\textbf{Freq.} \\ (\si{\hertz})}} &
    {\thead{\textbf{Latency} \\ (\si{\milli\second})}} & 
    {\thead{\textbf{Energy} \\ (\si{\milli\joule})}}
\\
\hline

    Cardio & SVM~\cite{Mubarik:MICRO:2020:printedml} & 
    90.0 & 15.1 & 57.4 & 13 & 75 & 4.31
\\
    Cardio & SVM~\cite{Armeniakos:TCAD2023:cross}$^\star$ & 
    89.0 & 17.0 & 48.9 & 13 & 75 & 3.67
\\
    Cardio & MLP~\cite{Armeniakos:TC2023:codesign}$^\star$ & 
    87.0 & 6.1 & 20.8 & 5 & 200 & 4.16 
\\
    Cardio & \oursTable{Ours} & 
    93.4 & 17.1 & 17.6 & 38 & 78 & 1.373

\\
\hline

    Derm. & SVM~\cite{Mubarik:MICRO:2020:printedml} & 
    97.2 & 60.4 & 182.9 & 8 & 120 & 21.95
\\
    Derm. & \oursTable{Ours} & 
    98.6 & 13.9 & 14.3 & 38 & 156 & 2.231

\\
\hline

    PD & SVM~\cite{Mubarik:MICRO:2020:printedml} & 
    97.8 & 123.8 & 364.4 & 4 & 250 & 91.1
\\
    PD & SVM~\cite{Armeniakos:TCAD2023:cross}$^\star$ & 
    97.0 & 97.0 & 183.7 & 4 & 250 & 45.92 
\\
    PD & MLP~\cite{Armeniakos:TC2023:codesign}$^\star$ &
    93.0 & 32.7 & 99.2 & 4 & 250 & 24.8 
\\
    PD & \oursTable{Ours} & 
    93.1 & 22.9 & 22.9 & 35 & 280 & 6.41

\\
\hline

    RW & SVM~\cite{Mubarik:MICRO:2020:printedml} & 
    57.0 & 23.5 & 92.8 & 15 & 66 & 6.12
\\
    RW & SVM~\cite{Armeniakos:TCAD2023:cross}$^\star$ & 
    56.0 & 11.7 & 21.3 & 15 & 66 & 1.41
\\
    RW & MLP~\cite{Armeniakos:TC2023:codesign}$^\star$ & 
    56.0 & 1.1 & 3.9 & 5 & 200 & 0.79
\\
    RW & \oursTable{Ours} & 
    64 & 6.2 & 6.7 & 42 & 144 & 0.965

\\
\hline

    WW & SVM~\cite{Mubarik:MICRO:2020:printedml} & 
    53.0 & 28.3 & 112.4 & 17 & 60 & 6.74
\\
    WW & SVM~\cite{Armeniakos:TCAD2023:cross}$^\star$ & 
    52.0 & 11.0 & 34.7 & 17 & 60 & 2.08 
\\
    WW & MLP~\cite{Armeniakos:TC2023:codesign}$^\star$ & 
    53.0 & 6.5 & 21.3 & 5 & 200 & 4.26 
\\
    WW & \oursTable{Ours} & 
    56 & 6 & 6.4 & 34 & 203 & 1.299

\\ 
\hline
\end{tabular}
\begin{tablenotes}[flushleft]\footnotesize
    \item[]$^\dagger$Derm.: Dermatology, PD: PenDigits, RW: RedWine, WW: WhiteWine 
    \item[]$^\star$Approximate model.
\end{tablenotes}
\end{threeparttable}
\vspace{-4ex}
\end{table}

\textbf{Results:}
\cref{tab:results} presents the results of our comparison against the state of the art, in terms of accuracy area, power, frequency, latency and energy.
Overall, our sequential \glspl{svm} feature the most favorable accuracy-energy trade-off among related approaches.
Across all datasets, we achieve higher accuracy by $2.02\%$, $3.13\%$, and $4.38\%$ on average, compared to each state-of-the-art technique~\cite{Mubarik:MICRO:2020:printedml}, \cite{Armeniakos:TCAD2023:cross}, and \cite{Armeniakos:TC2023:codesign}, respectively.
An exception can be found for the PenDigits dataset, where the state of the art employs a larger number of support vectors to retain accuracy, but incurs unrealistic hardware overheads..
Our \glspl{svm}, despite showing small area overheads in some cases, compared to related works, manage to stay within acceptable area ranges, satisfying the constraints of typical printed applications~\cite{Armeniakos:TCAD2023:cross}.
Additionally, our peak power consumption is \SI{22.9}{\milli\watt} and the average \SI{13.58}{\milli\watt}, which enables \emph{all} our designs to be powered by existing printed batteries (e.g., Molex \SI{30}{\milli\watt}).
In contrast, only $4$ designs of the state of the art can be powered by an existing printed power source.
Importantly, our circuits achieve an average energy consumption of only \SI{2.46}{\milli\joule}, which corresponds to a $10.6\times$ improvement over~\cite{Mubarik:MICRO:2020:printedml}, $5.4\times$ over~\cite{Armeniakos:TCAD2023:cross}, and $3.46\times$ over~\cite{Armeniakos:TC2023:codesign}.
Our circuits feature the best energy consumption in all but one dataset (RedWine), where the difference is only \SI{0.17}{\milli\joule}, but our accuracy is significantly better in that case ($8\%$ higher).
With an average energy improvement of $6.5\times$, our \glspl{svm} can successfully boost battery life in printed applications.

\section{Conclusion}
\label{sec:conclusion}
In this work, we propose highly-accurate and energy-efficient sequential \glspl{svm}, tailored for printed technology.
Our bespoke designs utilize the \gls{ovr} algorithm and perform classification over multiple cycles, limiting hardware overheads.
Compared to parallel alternatives from the state of the art, our circuits deliver higher accuracy, with an energy improvement of $6.5\times$, whilst satisfying the stringent area and power constraints of \gls{pe}.
Thus, our proposed designs can be powered by existing printed batteries (e.g., Molex \SI{30}{\milli\watt}), and even prolong their battery lifetime in printed applications.

\section*{Acknowledgment}
This work is funded by the H.F.R.I call “Basic research Financing (Horizontal support of all Sciences)” under the National Recovery and Resilience Plan “Greece 2.0” (H.F.R.I. Project Number: 17048).


\end{document}